\documentclass[10pt,letterpaper,twocolumn]{article} 

\usepackage{mathptmx}
\usepackage[utf8]{inputenc} 
\usepackage{amssymb,amsmath}
\usepackage[font=small,labelfont=bf]{caption}
\usepackage{color}
\usepackage[usenames,dvipsnames]{xcolor}

\usepackage{geometry}
\usepackage{graphicx,subcaption} 

\usepackage{epstopdf}

\usepackage{booktabs} 
\usepackage{array} 
\usepackage{paralist} 
\usepackage{verbatim} 

\usepackage{fancyhdr} 
\usepackage{sectsty}
\allsectionsfont{\sffamily\mdseries\upshape} 

\usepackage[nottoc,notlof,notlot]{tocbibind} 
\usepackage[titles,subfigure]{tocloft} 

\usepackage{setspace}
\usepackage{hyperref}

\title{"Temporal-Needle": A view and appearance invariant video descriptor}
\author{Michal Yarom \and Michal Irani}
\date{The Weizmann Institute of Science, Israel} 
\begin{document}
\maketitle

\begin{abstract}
\normalsize
\textit{The ability to detect similar actions across videos can be very useful for real-world applications in many fields. However, this task is still challenging for existing systems, since videos that present the same action, can be taken from significantly different viewing directions, performed by different actors and backgrounds and under various video qualities. Video descriptors play a significant role in these systems.}

\textit{In this work we propose the "temporal-needle" descriptor which captures the dynamic behavior, while being invariant to viewpoint and appearance. The descriptor is computed using multi temporal scales of the video and by computing self-similarity for every patch through time in every temporal scale. The descriptor is computed for every pixel in the video. However, to find similar actions across videos, we consider only a small subset of the descriptors - the statistical significant descriptors. This allow us to find good correspondences across videos more efficiently. Using the descriptor, we were able to detect the same behavior across videos in a variety of scenarios.}

\textit{We demonstrate the use of the descriptor in tasks such as temporal and spatial alignment, action detection and even show its potential in unsupervised video clustering into categories.
In this work we handled only videos taken with stationary cameras, but the descriptor can be extended to handle moving camera as well.}

\end{abstract}



\section{Introduction}
Action analysis has drawn significant attention of the computer vision community. Although there has been progress in the past two decades, it is still considered a hard challenge, especially in unconstrained videos. The interest in the topic is motivated by the potential of many applications based on automatic video analysis, ranging from video retrieval, surveillance systems, machine-human interaction and sports video analysis.

The problem we addressed in our work is to develop a descriptor for action detection, which will allow detecting the same action in different videos recorded from different view points, possibly at different times and even by different sensing modalities. We considered two cases: a simple case where the videos capture the same scene recorded simultaneously from different viewing directions, and the general case where they capture different scenes, i.e. the same action performed by different actors with different backgrounds.

This gives the motivation to develop a descriptor which will capture the dynamics in the video, while being invariant to appearance, view-point, scale and insensitive to small temporal variations.

Various approaches have been proposed over the years for action detection and recognition: ranging from high level representation of shapes and silhouettes to low-level appearance and motion estimation. 
Several early attempts used silhouettes to extract human motion such as \cite{ActionsAsSpaceTimeShapes_iccv05, silhouettes:1, silhouettes:2,silhouettes:3}. 

A recent work by Ben-Artzi et al \cite{Ben-Artzi:1} propose to create a "Motion Barcode" for every pixel, such that it captures the exiting/non-exiting motion in that pixel over time. Their method can determine if two videos presents the same event even if they are captured from significant different viewing directions. However, their method is limited to videos that capture simultaneously the same dynamic scene, and they rely on background-foreground segmentation.

In a more recent work by the same authors \cite{Ben-Artzi:2}, they combine dynamic silhouette methods with their temporal signature, to estimate the epipolar geometry between the cameras that capture the same event. The temporal signature is similar to the Motion Barcode, but instead of computing the signature for every pixel, they compute a signature for epipolar lines. This work is still limited to the same scene and relies on extracting silhouettes.

In the recent years, most works focused on local low level representations, and are briefly reviewed below. For more comprehensive survey the reader is referred to \cite{Poppe:1,Survey:1}.

One type of low level representation is done by first finding Space-Time Interest Points (STIP) \cite{Laptev:2}. The local information in these points is captured using one of several descriptors, such as HoG, HoF, SIFT \cite{hog:1,hof:1,SIFT:1} or a 3D modified version for them (e.g. \cite{hog3D:1}). Then the video is represented by a Bag of Words technique \cite{bagOfWords:1}. This approach has proven effective in action recognition on challenging datasets (e.g. in \cite{bagOfWords:2}). 

In \cite{Michal:3} a method for alignment of video sequences from different modalities is proposed. They demonstrate that by tracking the space-time interest points, they can estimate the temporal and spatial sequence to sequence alignment using trajectories.
However, the main limitation of the STIP approach is that it relies on finding a suitable amount of space-time interest points. Videos with subtle motion will not provide enough corresponding interests points across the videos. On the other hand, videos with large motions (e.g. videos of waves in the sea) will provide too many interest points, making it difficult to find reliable matchings.

In \cite{Michal:4} a direct approach for sequence alignment is proposed, based on maximizing local space-time correlations. The algorithm is applied directly on the space-time intensity information without finding space-time interest points. They were able to align challenging sequences with different appearance and time fluctuations. However, this approach is restricted to 2D parametric transformations, and can not be applied to sequences taken from different viewpoints.

Shechtman and Irani \cite{Michal:1} propose a self-similarity descriptor that correlate space-time local patches over different locations in space and time. The descriptor is invariant to color, texture and to small scale variations. They have proved its utility for action detection between videos with different appearances. However, if there is large scale variations (e.g, due to zoom difference between the videos, such as in Fig.~\ref{fig:zoomAlign}), or large viewpoint variations, the self-similarity descriptor will fail to detect the same action.

Junejo et al \cite{Laptev:1} have shown that the temporal self-similarity matrices (SSM) of an action seen from different viewpoints are very similar. The temporal SSM can be used with different local descriptors. This method was successfully used for cross-view action recognition. However, it is restricted to a single action within the field of view.

Kliper-Gross et al \cite{Orit:1} developed a Motion Interchange Pattern (MIP) descriptor that creates a signature at every pixel and triplet of frames in the video (the previous, current and next frame). It encodes the motion by comparing the patch centered in the pixel coordinate at the current frame, to 8  patches in the previous and next frames. This is an extension of the Local Trinary Triplet (LTP) descriptor \cite{Wolf:1}. The LTP descriptor only compares the patches, in the previous and next frames, on the same location relative to the patch in the current frame, i.e. 8 comparisons instead of 64 in the MIP descriptor. By Combining the MIP descriptor with a standard bag of words technique, it achieved impressive results of action recognition on challenging benchmarks. However, their method is also restricted to a single action in the field of view.
\\
\\In this work we introduce a new space-time descriptor, the "temporal-needle", which captures dynamics. The descriptor is invariant to changes in appearance, view-point and geometric transformations.
The key to our work is creating a signature by computing self-similarity over time at the same pixel location, at multiple temporal scales. It creates a signature of local repetitive behavior at every point in space and time.

Inspired by the self-similarity and the MIP descriptors, we compute the sum of squared differences (SSD) between small patches. However, in our descriptor, the SSD is computed between a small spatial patch around a point in the current frame, to the patches centered at the same spatial point in the neighboring frames. Unlike the MIP descriptor, we compute the descriptor for a larger number of frames, and at multiple temporal scales. By increasing the temporal support we can better represent the motion patterns of the action. By computing the descriptor in multi temporal scales of the video, allows for subtle temporal variations in speed of the same action.

The descriptor will be described in more details in Sec.~\ref{sec:overview} and Sec.~\ref{sec:descriptor}. We describe an efficient method for finding good correspondences across videos in Sec.~\ref{goodCorrespondences}. We test the applicability of the temporal needle descriptor for a variety of tasks. Sec.~\ref{seqToSeq} demonstrates its use for temporal and spatial alignment of video sequences for a large variety of scenarios: videos taken with different types of sensors, videos with wide baseline, videos with non-rigid motion and videos with significant zoom difference. Sec.~\ref{sec:actionDetection} shows its use for action detection; we were able to detect the same action in videos of a real sports games. Sec.~\ref{sec:videoClustering} presents its potential use for unsupervised video clustering.

\section{Overview of Our Approach} \label{sec:overview}
Structure from motion (SfM) \cite{HartleyZisserman:1} and photo tourism \cite{modeling3D:1} recover camera parameters and the fundamental matrix between images, by finding correspondences between images of the same scene based on their appearance (usually by using feature detectors such as SIFT \cite{SIFT:1} and SURF \cite{SURF:1}).

We extend these ideas to videos. We want to compare and find correspondences between two videos $V_1(x,y,t)$ and $V_2(x,y,t)$ under the assumption that they capture the same (or similar) action. The action takes place in the 4D space-time world $(X,Y,Z,T)$. Videos $V_1$ and $V_2$ are the projections of the 4D action into a 3D coordinate systems $(x,y,t)$, defined by the internal and external parameters of the video cameras.

For simplicity, let's assume that cameras are stationary and synchronized in time. Suppose that the videos show a dancer, and that his hand passes at the global coordinate $(X,Y,Z)$ at some discrete times $t_1, t_2, \dots, t_n$. Since the camera parameters are fixed, we expect that the dancer's hand to be projected on some point $(x_1,y_1)$ in $V_1$ and on a point $(x_2,y_2)$ in $V_2$, at the same times $t_1,t_2,\dots, t_n$. Namely, there will be strong self-similarity between the patch around $(x_1,y_1)$ in $V_1$ across the frames at times $t_1,t_2,\dots,t_n$ in $V_1$. Similarly, there will be strong self-similarity between the patch around $(x_2,y_2)$ in $V_2$ across the frames at times $t_1,t_2,\dots,t_n$ in $V_2$. Therefore, by creating a signature of self-similarities for each spatial location over time in the videos $V_1$ and $V_2$, we will be able to recognize that $(x_1,y_1)$ and $(x_2,y_2)$ present the same behavior, hence corresponding dynamic points.

This gives the motivation why the temporal needle descriptor is view-invariant. In Sec.~\ref{sec:descriptor} we explain in details why the descriptor is view-invariant and appearance-invariant.

Similarly, we claim that the descriptor is invariant to scale. Let's assume that the dancer is captured from the same direction by two videos with zoom ratio of 1:$a$. As assumed before, the hand of the dancer passes at time $t_1$ through point $(x_1,y_1)$ in $V_1$ and through point $(x_2,y_2)$ in $V_2$. Let $b$ denote the number of pixels the hand moves from $(x_1,y_1)$ in $V_1$ between $t_1$ to the next frame. Due to the zoom ratio between the videos, the number of pixels the hand moves from $(x_2,y_2)$ in $V_2$ will be $a \cdot b$ at the same time. The number of pixels the hand moves in each video is different, but eventually in time $t_2$ it returns back to point $(x_1,y_1)$ in $V_1$ and to point $(x_2,y_2)$ in $V_2$. 

Our descriptor does not estimate the motion, but instead it measures the self-similarity of a fixed spatial point through time. There will be strong self-similarity between the patch centered at $(x_1,y_1)$ in $V_1$ at time $t_1$ to patches at the same spatial location in every frame the hand is passes through point $(x_1,y_1)$ (in our example in times $t_1,t_2,\dots,t_n$). Similarly, there will be strong self-similarity between the patch centered at $(x_2,y_2)$ in $V_2$ at the same times $t_1,t_2,\dots,t_n$. Thus, by using our descriptor, we will be able to recognize that points $(x_1,y_1)$ and $(x_2,y_2)$ are corresponding dynamic points, despite the different zoom.

We compute the SSD (sum of squared differences) of a small spatial patch in the current frame with patches, located at the same spatial location, in its neighboring frames. We measure the self-similarity in more than one temporal scale, by scaling down the video in time, thus generating down-sampled versions of the video $\{V^s\}$. 
We were inspired to use temporal multi-scale versions of the video in our temporal needle descriptor by the spatial Needle descriptor proposed by \cite{Or:1}. 
We have found that adding temporal scales is better than increasing the temporal length because it captures more significant parts of the action, while being insensitive to variation in the speed of the action, as explained in Sec.~\ref{sec:descriptor}.

Fig.~\ref{fig:tennisServe} illustrates the properties of the descriptor. It shows two corresponding descriptors that were extracted from videos of a tennis serve. The descriptors of the corresponding points are very similar, although the action is performed by different tennis players and different backgrounds, from significantly different viewpoints, and there are slight speed variations between the players.

\section{The Temporal-Needle Descriptor}\label{sec:descriptor}
In this section we introduce our multi scale temporal-needle descriptor, which is computed for every pixel in the video.
Let $V_1(x,y,t)$ and $V_2(x,y,t)$ be two videos that capture the same action. We would like to find good matching points between the two videos. Let $p_1 = (x_1,y_1,t_1)$ be a point in the first video $V_1$, and suppose that its matching point in the second video $V_2$ is located at $p_2=(x_2,y_2,t_2)$.
Although the videos present the same action, they can be very different: the actors, their clothes, their backgrounds, illumination, the viewpoint and zoom of the cameras, all these can change between the videos.
Therefore, we are interested in finding good correspondences based on the dynamic behavior presented in the videos.

The temporal-needle signature, $d(x,y,t)$ for a point located at $(x,y,t)$ is computed as follows: the video $V$ is downscaled temporally, to generate a multi scale temporal pyramid $\{V^s\}$, by blurring with a Gaussian LPF along the temporal dimension, and then sub-sampling the video $V$ in time. For every patch $p$ and every scale $s$ ($s = 1,\frac{1}{2}, \frac{1}{4},\dots$) in the video pyramid $V^s$, we compute: the sum of squared differences (SSD) between the patch $p$ centered at $(x,y,t)$ with the patches located in the same spatial coordinates $(x,y)$ in the $\Gamma$ previous and $\Gamma$ next frames in that scale. This creates a vector of length $2\Gamma$, per point $(x,y,t)$ per scale $s$:
\begin{equation}
\begin{aligned}
d^s(x,y,t) = & [\|p(x,y,t,s) - p(x,y,t-\Gamma,s)\|^2,\dots, \\
& \|p(x,y,t,s) - p(x,y,t+r,s)\|^2,\dots, \\
& \|p(x,y,t,s) - p(x,y,t+\Gamma,s)\|^2]
\end{aligned}
\end{equation}
Where $-\Gamma \leq r \leq \Gamma$. The descriptor is a vector of length $2\Gamma$ (after omitting the 0 value at its center when $r=0$).

The temporal-needle descriptor for point $(x,y,t)$ in the video $V$ is obtained by concatenating the self-similarities corresponding to that point from all scales of the video pyramid $V^s$.
\begin{equation}
d(x,y,t) = [d^1(x,y,t), d^{1/2}(x,y,t/2), d^{1/4}(x,y,t/4),\dots]
\end{equation}
Finally, the descriptor is normalized, so that the sum of all its entries are 1:
\begin{equation}
d(x,y,t) = \frac{d(x,y,t)}{max(Sum(d(x,y,t)), Sum_{noise})}
\end{equation}
In the experiments we present, we sub-sampled the video by a factor of 0.5. The normalization is done by choosing the maximum between the sum of the elements in the descriptor and $Sum_{noise}$. $Sum_{noise}$ is equal to the descriptor's length multiplied by a constant that represents the estimated noise variance. In our experiments the noise variance was set to be the $30^{th}$ percentile of all entry values, taken from all descriptors computed for that video.

\begin{figure}[h!]
	\centering
	\begin{subfigure}[b]{0.4\textwidth}
		\centering
		\includegraphics[width=\textwidth]{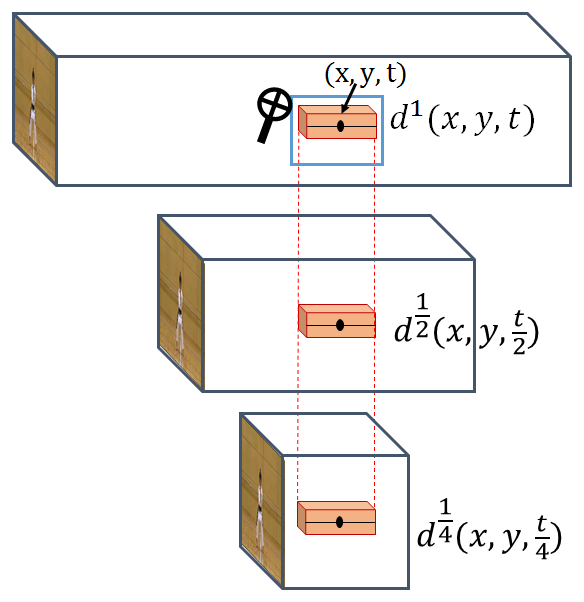}
		\caption{}
		\label{fig:descriptorA}
	\end{subfigure}
	\hfill
	\begin{subfigure}[b]{0.4\textwidth}
		\centering
		\includegraphics[width=\textwidth]{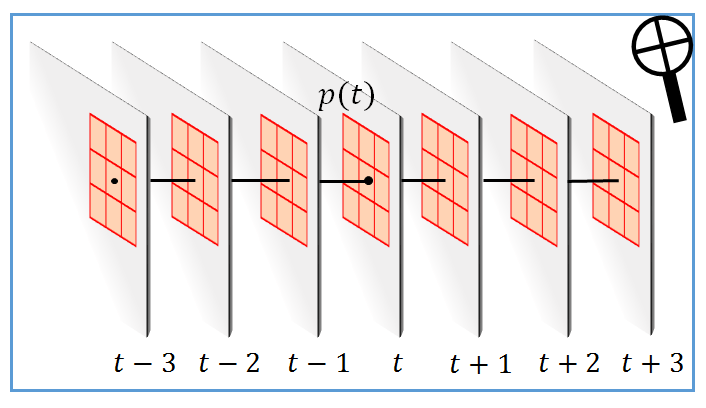}
		\caption{}
		\label{fig:descriptorB}
	\end{subfigure}
	\caption{(a) The Temporal Needle descriptor with 3 temporal levels. (b) The computation of the descriptor in every temporal scale for $\Gamma=3$ and patch size 3x3. We compute the SSD between the patch centered at $(x,y,t)$ with the patches at the same spatial position in the $\Gamma$ previous and $\Gamma$ next frames.}
	\label{fig:descriptor}
\end{figure}

Fig.~\ref{fig:descriptor} illustrates how we compute the descriptor in every temporal scale for $\Gamma = 3$.

The temporal needle descriptor has two important properties which lead to finding good correspondences across videos.
\subsection{Large temporal context with flexibility } 
Increasing the temporal support ($2\Gamma$+1) of the descriptor will enable to represent longer and more meaningful repetitive behaviors. On the other hand, the self-similarity pattern will need to be the same for a larger number of frames across videos, which will not allow flexibility in the speed of the action. We next show that adding more temporal scales achieves these two goals: increasing the temporal context, while allowing for flexibility in the speed of the action. 

Let's assume that videos $V_1$ and $V_2$ capture two different people performing similar actions. Let $(x_1,y_1,t_1)$ in $V_1$ and $(x_2,y_2,t_2)$ in $V_2$ be two corresponding points. Denote by $\lambda = 2\Gamma +1$ the temporal support of the temporal needle descriptor.
\\Let's assume that the temporal needle descriptors of these corresponding points are computed with $L$ temporal scales, i.e. $d_{V_1}(x_1,y_1,t_1) = [d^1_{V_1}(x_1,y_1,t_1), \dots , d^{s_L}_{V_1}(x_1,y_1,t_1/s_L)]$ and \\$d_{V_2}(x_2,y_2,t_2) = [d^1_{V_2}(x_2,y_2,t_2), \dots , d^{s_L}_{V_2}(x_2,y_2,t_2/s_L)]$. The descriptors $d^{s_l}_{V_1}(x_1,y_1,t_1/s_l)$ and $d^{s_l}_{V_2}(x_2,y_2,t_2/s_l)$ are computed in a down scaled versions of $V_1$ and $V_2$ by a factor of $s_l$. Let $\tau_1$ and $\tau_2$ be the temporal windows (of length $\lambda$) captured by the descriptors in the coarsest temporal scale of $V_1$ and $V_2$. Let $T_1$ and $T_2$ be the temporal windows in the original temporal scale of $V_1$ and $V_2$ which correspond to $\tau_1$ and $\tau_2$ in the coarse scale, namely down-scaling $T_1$ and $T_2$ temporally by a factor of $s_L$ results in $\tau_1$ and $\tau_2$, respectively: $T_1 \downarrow_{1/s_L} = \tau_1$ and $T_2 \downarrow_{1/s_L} = \tau_2$. Their length $|T_1| = |T_2| = \Lambda > \lambda$. 

\begin{figure}[h!]
	\centering
	\includegraphics[width=0.5\textwidth]{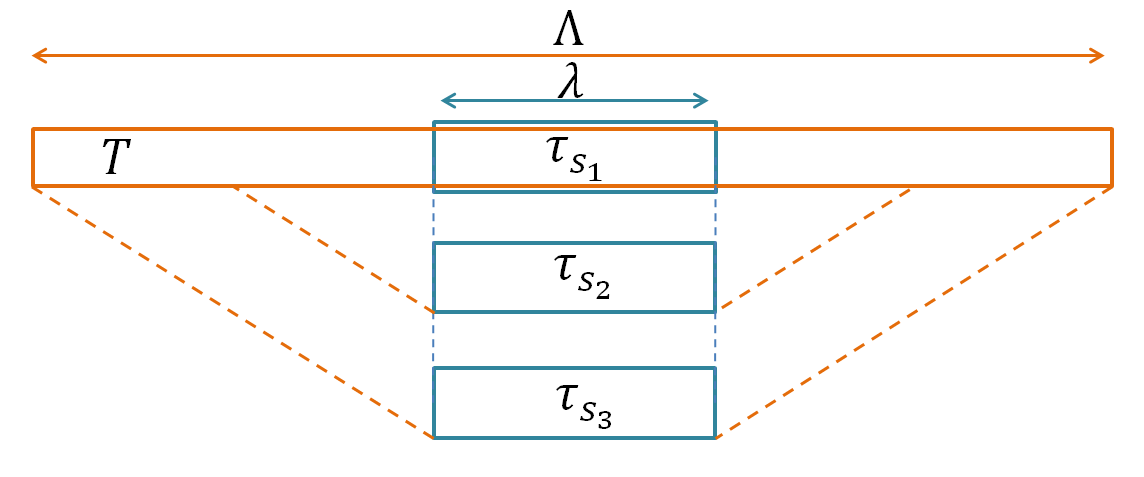}
	\caption{illustration of the correspondences between temporal windows in a scale-down version of the video, to windows in the original temporal scale. $\tau_{s_1}$, $\tau_{s_2}$ and $\tau_{s_3}$ are temporal windows, of length $\lambda$, in 3 temporal scales of the video. The window $\tau_{s_3}$, in the coarsest temporal scale of the video, corresponds to the window $T$, in the original temporal scale of the video. Similarly, the window $\tau_{s_2}$ corresponds to a smaller window in the original temporal scale of the video.}
	\label{fig:temporalMisalignmentNeedle}
\end{figure}

Fig.~\ref{fig:temporalMisalignmentNeedle} illustrate the correspondences between windows in a temporal down-scaled versions of the video, to the windows in the original temporal scale. $\tau_{s_1}$, $\tau_{s_2}$ and $\tau_{s_3}$ are temporal windows, of length $\lambda$, in 3 temporal scales of the video. The window $\tau_{s_3}$ in the coarsest temporal scale of the video, corresponds to the window $T$ in the original temporal scale of the video. Similarly, the window $\tau_{s_2}$ corresponds to a smaller window in the original temporal scale of the video (marked by dashed lines).

\begin{figure}[h!]
	\centering
	\begin{subfigure}[b]{0.55\textwidth}
		\centering
		\includegraphics[width=\textwidth]{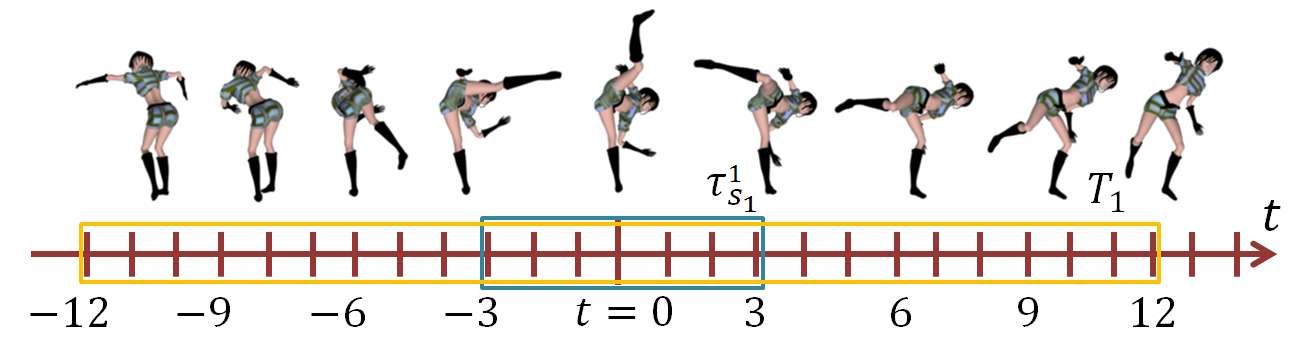}
		\caption{Video $V_1$}
		\label{fig:tempMis_V1}
	\end{subfigure}
	\hfill
	\begin{subfigure}[b]{0.55\textwidth}
		\centering
		\includegraphics[width=\textwidth]{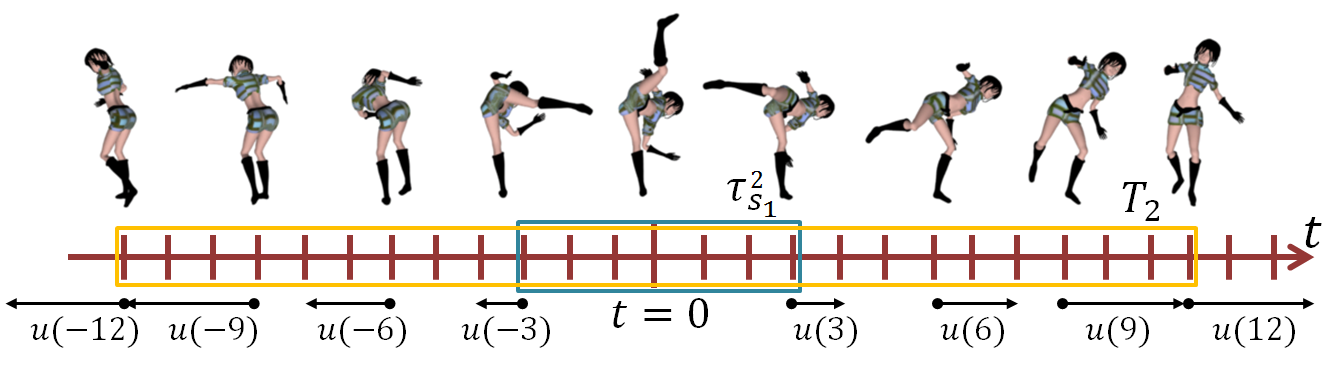}
		\caption{Video $V_2$}
		\label{fig:tempMis_V2}
	\end{subfigure}
	\caption{\textbf{Example of temporal misalignment between videos presenting the same action in different speed.} $V_1$ and $V_2$ capture a similar action perfromed at different speeds. (a) and (b) show a few frames from the action in $V_1$ and in $V_2$. The center frame of the kick is in time $t=0$ in both videos. One can see that the kick in $V_2$ is faster than in $V_1$ by a speed ratio of $4/3$. The position of the girl in $V_2$ in times $t=9$ and $t=-9$ is the same as in $V_1$ in times $t=12$ and $t=-12$. The black arrows in (b) present the relative misalignment, $u(t)$, between the frames in $V_2$ to the frames in $V_1$.}
	\label{fig:temporalMisalignment}
\end{figure}

Assume we compute the self-similarity descriptor $d_{T_1}$ of window $T_1$ only at the finest scale, namely, the self-similarity of the central patch in the window $T_1$ to all other patches at the same spatial location in all other frames in $T_1$. Let $d_{T_2}$ be the self-similarity descriptor of window $T_2$. Let $u(t)$ denote the "misalignment" between $d_{T_1}$ and $d_{T_2}$.
These temporal deformations can occur, for example, if the action in one video is performed in different speed than in the second video, such as shown in Fig.~\ref{fig:temporalMisalignment}. In this example $V_1$ and $V_2$ present a similar action performed at different speeds. One can notice that in $V_2$ the girl kicks faster than in $V_1$ (by a speed ratio of $4/3$). In times $t=9$ and $t=-9$ in $V_2$ the girl is in the same position as in times $t=12$ and $t=-12$ in $V_1$. The black arrows in Fig.~\ref{fig:temporalMisalignment}(b) illustrate a deformation between the frames in $T_2$ and $T_1$, namely how much a frame in $T_2$ needs to move relative to the "matching" frame in $T_1$.

To simplify the analysis, we assume that the size of these temporal deformations (misalignments) between $T_1$ and $T_2$ grow at most linearly with the distance from the center frame $t=0$, namely:
\begin{equation}\label{eq:temproalRatio}
|u(t)| = \alpha|t|
\end{equation}
for some scalar $\alpha > 0$. We next show that although the average temporal misalignment is large for the temporal supports $T_1$ and $T_2$, it is small for the corresponding temporal supports $\tau_1$ and $\tau_2$. Moreover, the average temporal misalignment is small for the entire temporal needle descriptors $d(x_1,y_1,t_1)$ and $d(x_2,y_2,t_2)$.

For simplicity, we perform the computation of the average temporal misalignment in the continuum. Under the assumption of Eq.~\ref{eq:temproalRatio}, the average temporal misalignment per frame in  $T_1$ and $T_2$ with temporal size $\Lambda$ is:
\begin{equation}\label{eq:avgMisalignmentBig}
\begin{aligned}
& AvgTempMisalignment(T_1,T_2)=
\\& \frac{1}{\Lambda} \int_{-\Lambda/2}^{\Lambda/2} |u(t)| dt =
\\& \frac{1}{\Lambda} \int_{-\Lambda/2}^{\Lambda/2} \alpha|t| dt
= \frac{\Lambda \alpha}{4}
\end{aligned}
\end{equation}
Thus, if we were to compute a self-similarity temporal descriptor for these two windows, $T_1$ and $T_2$, this would be the average "misalignment" between the entries of their descriptors $d_{T_1}$ and $d_{T_2}$. This, however, is not true for descriptors estimated on the down-scaled temporal windows $\tau_1$ and $\tau_2$.
We claim that the average temporal misalignment of descriptors estimated on  $\tau_1$ and $\tau_2$ will be significantly smaller. When two video are temporally scaled-down by a factor of $s_l$, their relative temporal misalignments become $s_l$-times smaller. In our case, $|u\downarrow_{1/s_l} (t)| = \frac{1}{s_l} |u(s_lt)| = \frac{\alpha}{s_l} |s_l t| = \alpha|t|$. Hence, the constant $\alpha$ remains the same in all temporal scales. Under the assumption of Eq.~\ref{eq:temproalRatio}, the average temporal misalignment per frame (hence also for descriptor entry) in  $\tau_1$ and $\tau_2$ with temporal size $\lambda$ is:
\begin{equation}\label{eq:avgMisalignmentSmall}
\begin{aligned}
& AvgTempMisalignment(\tau_1,\tau_2) = 
\\& \frac{1}{\lambda} \int_{-\lambda/2}^{\lambda/2} \alpha |t| dt
= \frac{\lambda \alpha}{4}
\end{aligned}
\end{equation}
Thus, the average "misalignment" between the entries of the self-similarity temporal descriptor for these two coarse windows is:
\small
$AvgTempMisalignment(d^{s_L}(x_1,y_1,t_1/s_L),d^{s_L}(x_2,y_2,t_2/s_L)) = \frac{\lambda \alpha}{4}$. 
\normalsize
\\This is true for the descriptors in every temporal scale (the average  "misalignment" is independent of the scale $s_l$). Therefore, the average temporal misalignment per entry between the  temporal needle descriptors $d(x_1,y_1,t_1)$ and $d(x_2,y_2,t_2)$ is:
\begin{equation}\label{eq:tempNeedleAvgMisalign}
\small
\begin{aligned}
&AvgTempMisalignment(d(x_1,y_1,t_1),d(x_2,y_2,t_2)) = 
\\&\frac{1}{L} \sum_{l=1}^L AvgTempMisalignment(d^{s_l}(x_1,y_1,t_1/s_l),d^{s_l}(x_2,y_2,t_2/s_l))
\\& = \frac{\lambda \alpha}{4}
\end{aligned}
\end{equation}
For example, let's assume that the speed of the action in video $V_2$ is faster than the speed of the action in video $V_1$ by a factor of $\beta \geq 1$, then $\alpha = |1 - \beta|$. In most of our experiments the descriptor's temporal length is $\lambda=7$ (i.e., radius 3), and we used $L=3$ temporal scales resulting in temporal context $\Lambda = 25$ (i.e., radius 12). For speed ratio $\beta = 1.25$, the average temporal misalignment between the entries of the self-similarity temporal descriptors for the windows $T_1$ and $T_2$ is $AvgTempMisalignment(T_1,T_2) \approx 1.6$ frames. Whereas, the average temporal misalignment per entry between the temporal needle descriptors $d(x_1,y_1,t_1)$ and $d(x_2,y_2,t_2)$ is $AvgTempMisalignment(d(x_1,y_1,t_1),d(x_2,y_2,t_2)) \approx 0.4$ frames. Namely, 4 times smaller. 

Moreover, the maximum misalignment between entries in the temporal self-similarity descriptors $d_{T_1}$ and $d_{T_2}$ of the windows $T_1$ and $T_2$, is $maxMisalignment(T_1,T_2) = 3$ frames. In fact, for 75\% of the entries in $d_{T_1}$ and $d_{T_2}$, the misalignment is larger than one frame. Whereas, the temporal needle descriptors, although they capture information from the same temporal context, the maximum misalignment between them is $maxMisalignment(d(x_1,y_1,t_1),d(x_2,y_2,t_2)) = 0.75$ frames. The small misalignments lead to high similarity between the two temporal needles. Therefore, adding more temporal scales to the needle is equivalent to increasing the temporal context, with the advantage of being insensitive to small speed variations.


\subsection{Appearance-invariance \& View-invariance}
We will only demonstrate the properties of the descriptor in a single temporal scale as it applies to every temporal scale, it therefore applies to the entire descriptor.

\begin{figure}[h!]
	\centering
	\includegraphics[width=0.55\textwidth]{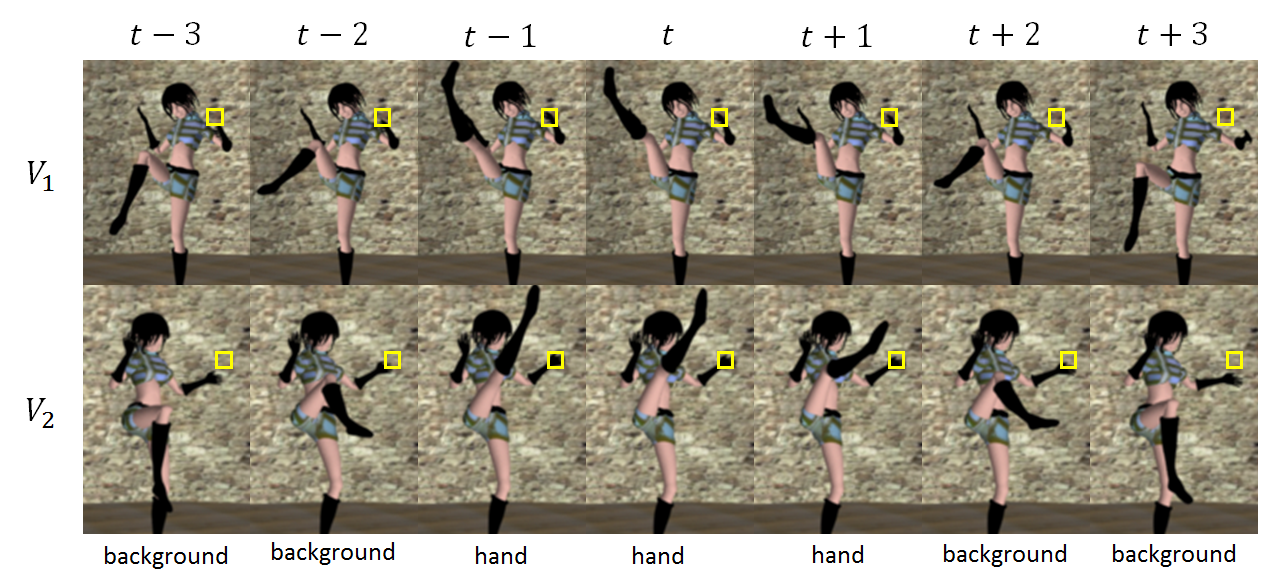}
	\caption{An example of the same action captured from two different viewpoints. Althogh the videos are taken from different views, the patch marked by a yellow rectangle captures the girl's hand in frames $t-1,t$ and $t+1$. In the rest of the frames the patch captures the background (note that in $V_1$ the background is darker than in $V_2$).}
	\label{fig:kungFuKick}
\end{figure}

Suppose that the two videos $V_1$ and $V_2$ capture the same action from different view points (e.g., see Fig.~\ref{fig:kungFuKick}). The action takes place in a 4D space-time world and videos $V_1(x,y,t)$ and $V_2(x,y,t)$ are the 3D projections of the action. We assume that the hand of the actor passes through some coordinate $(X,Y,Z)$ in discrete times $t_1,t_2,\dots,t_n$. Since the positions of the cameras are fixed this point is projected to some point $(x_1,y_1)$ in $V_1$ and to some point $(x_2,y_2)$ in $V_2$. 

For example in Fig.\ref{fig:kungFuKick}, the patch centered at the point $(x_1,y_1)$ in $V_1$ and the patch centered at the point $(x_2,y_2)$ in $V_2$ capture the girl's hand in frames $t-1,t,t+1$, and background in frames $t-3,t-2,t+2,t+3$. When we compute the descriptor $d^1(x_1,y_1,t)$ in $V_1$ and the descriptor $d^1(x_2,y_2,t)$ in $V_2$ with $\Gamma=3$, we measure the SSD of the patch located in $(x_1,y_1)$ (or $(x_2,y_2)$) in frame $t$ and the patches in the same spatial location in 3 previous frames and 3 next frames.

The SSD between the patch in frame $t$ to the patches in frames $t-1$ and $t+1$ will be approximately 0 (the patches capture the girl's hand). However, since $V_1$ and $V_2$ are taken from different viewpoints, in frames $t-3, t-2, t+2$ and $t+3$ the patches capture possibly different backgrounds. Therefore, for $V_1$ the SSD between the patch in frame $t$ to the patches in frames $t-3, t-2, t+2$ and $t+3$ will have some value $\alpha > 0$. Similarly, for $V_2$ the SSD between the patch in frame $t$ to the patches in frames $t-3, t-2, t+2$ and $t+3$ will have some different value $\beta > 0$ (usually $\beta \neq \alpha$). Thus, $d^1(x_1,y_1,t) = \begin{bmatrix} \alpha & \alpha & 0 & 0 & \alpha & \alpha \end{bmatrix}$ and $d^1(x_2,y_2,t) = \begin{bmatrix} \beta & \beta & 0 & 0 & \beta & \beta \end{bmatrix}$.

Then we normalize the descriptor by dividing each entry by the sum of the entries in the descriptor (we assume that this sum is larger than $Sum_{noise}$), and the normalized descriptors will be the same $d^1(x_1,y_1,t) = d^1(x_2,y_2,t) = \begin{bmatrix} \frac{1}{4} & \frac{1}{4} & 0 & 0 & \frac{1}{4} & \frac{1}{4}\end{bmatrix}$.
\\The normalization compensates for variations in the SSD of the patch in frame $t$ to patches with different appearance in its neighboring frames.

Similarly, we claim that the descriptor is appearance-invariant. Suppose that the appearance of the girl in $V_1$ is different than in $V_2$ (e.g., by capturing the scene with different types of sensors). The descriptor is computed by measuring the self-similarity of a patch around a point through time, i.e. measuring how much a patch around a point is similar to itself in different frames. Although the girls in the videos have different appearance, we only compare each video to itself. As in the example above, the final step of normalizing the descriptor makes it invariant to appearance.

\begin{figure*}[h!t]
	\centering
	\includegraphics[width=1\textwidth]{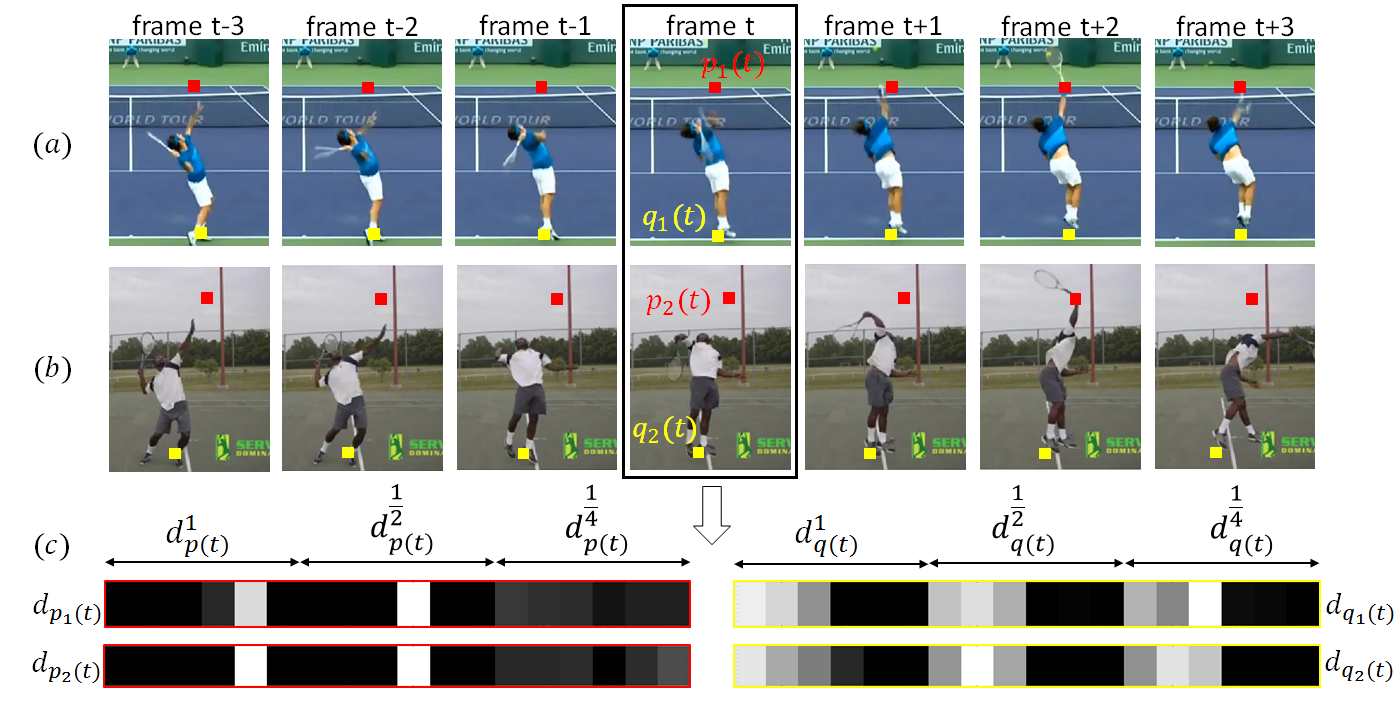}
	\caption{The temporal-needle of corresponding space-time points in two videos that present the same action. (a) presents 7 frames from a tennis serve by Roger Federer, we picked two points, $p_1(t)$ and $q_1(t)$, in the center frame $t$. (b) display 7 frames from a tennis serve by some other player and the points $p_2(t)$ and $q_2(t)$ are the corresponding points of $p_1(t)$ and $q_1(t)$, respectively. (c) display the descriptors of these 4 points. The descriptors were computed with patch size 3x3 (smaller than the rectangles presented in (a) and (b)), temporal radius of $\Gamma=3$ and 3 temporal scales. Although the videos were taken from very different view points and with different players and backgrounds, the descriptors of corresponding points are similar.}
	\label{fig:tennisServe}
\end{figure*}

Fig.~\ref{fig:tennisServe} shows an example of corresponding temporal-needles in two videos taken from different viewpoints of a tennis serve. As can be seen the tennis serve presented in both videos generate the similar local patterns over time, although it was performed by two different players, with different clothes and different backgrounds, and the videos were taken from different viewpoints. The first video was taken behind the player, while the second video was taken from the side. The temporal needle descriptor captures the local repetitive dynamics well, while being insensitive to differences in appearance and viewpoint.

\section{Good Video Correspondences} \label{goodCorrespondences}
In principal, the temporal needle descriptor can be computed for every pixel in the video, and then matched across videos. However, even short videos of a few seconds contain millions of pixels, most of them are background pixels, which are not particularly informative for matching. We next suggest an approach for focusing the correspondence estimation on only a small subset of informative descriptors.

\subsection{Extracting "informative" descriptors}
The distance between the descriptors of two corresponding video points must be small. However, this is not a sufficient condition to guarantee good correspondences. For example, points in the static background, will have a uniform zero descriptor, and hence will match well to any other static point. We therefore wish to match only dynamic points, and preferably those that produce reliable unambiguous matches. Thus, we would like to seek good matches between "informative" descriptors, namely, descriptors whose probability to appear at random is low.

Let $Q$ be a query video and $R$ be a reference video, we would like to match descriptors from $Q$ to $R$. We employ the notion of "saving in bits" and "informativeness" of descriptors, as defined in \cite{Michal:5, Michal:2}.

To find the "informative" descriptors in $Q$ we denote by $P_r(d|H)$ the probability of choosing the descriptor $d$ at random. We estimate $P_r(d|H)$ using a similar technique that was presented in \cite{Michal:2}. We generate a descriptor codebook $H$ as follows: first we sample a small portion (typically 1\%-5\%, as long as it more than 100,000 descriptors) of the descriptors from the two videos $Q$ and $R$. These descriptors are then clustered into a few hundreds clusters by applying K-means clustering. The centers (mean) of these clusters form the codebook words of H.

Descriptors that are very frequent in the videos, will be well represented in the codebook, whereas unique/rare descriptors will not be represented well in the codebook. Therefore, we approximate $P_r(d|H)$ by:
\begin{equation} \label{eq:informativeDescriptor}
P_r(d|H) = exp -\frac{|\Delta d(H)|^2}{2 \sigma ^2}
\end{equation}
where $\Delta d(H)$ denotes the distance between the descriptor $d$ and its closest word in the codebook $H$. If a descriptor $d$ is far from the codebook, it results in small estimated probability to appear at random, hence is flagged as an "informative" descriptor.

We further define the probability of finding a good match for descriptor $d$ in the reference video $R$ by $P_r(d|R)$. We use an approximation:
\begin{equation}\label{eq:nnDistance}
P_r(d|R) = exp -\frac{|\Delta d(R)|^2}{2 \sigma ^2}
\end{equation}
where $\Delta d(R)$ denotes the distance between the descriptor $d$ to its NN (nearest neighbor) descriptor in the reference video $R$.

We define the likelihood of a match for descriptor $d$ to be a "reliable" one as follows:
\begin{equation}\label{likelihoodRatio}
\text{Likelihood ratio}(d) = \frac{P_r(d|R)}{P_r(d|H)}
\end{equation}
Namely, the ratio between $P_r(d|R)$, the probability of finding a good match for descriptor $d$ in the reference video $R$, versus $P_r(d|H)$, the probability of finding the descriptor at random. Thus for example, if a descriptor $d \in Q$ has a good match in the reference video $R$ then $P_r(d|R)$ will be high. However, if $d$ is a trivial descriptor, then $P_r(d|H)$ will also be high, which results in an overall low likelihood ratio. On the other hand, if $d$ is informative, $P_r(d|H)$ will be low resulting in high likelihood ratio.

According to Shannon \cite{Shannon:1}, the entropy of a random variable $x$, namely $-p(x)log p(x)$, represents the number of bits required to code $x$. Therefore, taking the log of Eq.~\ref{likelihoodRatio} and discarding constants yields:
\begin{equation}
\text{"Saving in bits"}(d) = |\Delta d(H)|^2 - |\Delta d(R)|^2
\end{equation}
Thus, a descriptor whose distance $\Delta d(H)$ from the codebook $H$ is large (i.e., a rare descriptor), and found a good match in the reference video $R$ (i.e., the distance $\Delta d(R)$ is small), will result in high "saving in bits". This indicates a reliable match. It is not hard to see that if a descriptor is not informative or did not find a similar descriptor in the reference video, it will result in low "saving in bits", i.e., an unreliable match.
\\
\\In the next few sections (Sec.~\ref{seqToSeq}, \ref{sec:actionDetection}, \ref{sec:videoClustering}) we present several different applications of the temporal-needle descriptor. All the applications are based on finding reliable correspondences between the videos. These include Sequence-to-Sequence alignment (Sec.~\ref{seqToSeq}), Action detection (Sec.~\ref{sec:actionDetection}), and Video clustering (Sec.~\ref{sec:videoClustering}).

\section{Sequence to Sequence Alignment} \label{seqToSeq}
Let $V_1(x,y,t)$ and $V_2(x,y,t)$ be two videos capturing the same dynamic scene. Let $p_1=(x_1,y_1,t_1)$ be a point in the first video $V_1$ (namely, $p$ is a point in frame $t_1$ which is located in coordinates $(x_1,y_1)$ spatially), and let $p_2 = (x_2,y_2,t_2)$ be its matching point in the second video $V_2$. We assume that the cameras are stationary (they can also move jointly, as long as the parameters between the cameras are fixed) and the scene is dynamic.
\\We would like to find both the temporal and spatial alignment between $V_1$ and $V_2$. Correspondences between the videos both in space and time can be modeled with a small set of parameters, $T_{spatial}$ and $T_{temporal}$, and our goal is to find these parameters.

\subsection{Temporal Alignment} \label{tempAlignment}
Videos $V_1$ and $V_2$ can be misaligned temporally if the cameras have different frame rates (which results in scaling in time) and if the cameras are not synchronized (which results in an offset in time).
Therefore, we model the temporal transformation between the two sequences as a 1D affine transformation in time: $t_2 = rt_1 + \Delta t$. Where $r$ is the temporal scaling (e.g., the ratio between the frame rates of the videos), and $\Delta t$ is the frame shift between them (which is not necessarily an integer number of frames). In most cases $r$ is known, therefore we only compute the shift (although we can also compute $r$).
Computing the time shift is done in two steps: first we compute a course  integer frame shift, and then we refine it to a sub-frame shift.
\begin{enumerate}[(a)]
\item \textbf{Computing an integer frame shift - }Let $S_1$ and $S_2$ denote the informative descriptors detected in $V_1$ and $V_2$, respectively (see Sec.~\ref{goodCorrespondences}). We search for an integer $\Delta t'$ in the range $[\Delta t_{min}, \Delta t_{max}]$ that will maximize similarity of informative descriptors between frames $t_1\in V_1$ and the corresponding frame $t_2\in V_2$ such that $t_2 = r t_1 + \Delta t'$. For each $\Delta t'$ in the range we align the videos according to the current temporal shift. We consider only the frames that overlap between the two videos, as illustrated in Fig.~\ref{fig:integerFrameShift}. For every informative descriptor in $S_1$ we seek its NN (nearest neighbor) in its corresponding frame (according to the current shift). We measure the SSD between the descriptors. We repeat this process for the descriptors in $S_2$. We choose the integer frame shift $\Delta t'$ that minimizes the average error per descriptor.

\begin{figure}[h!]
	\centering
	\includegraphics[width=0.47\textwidth]{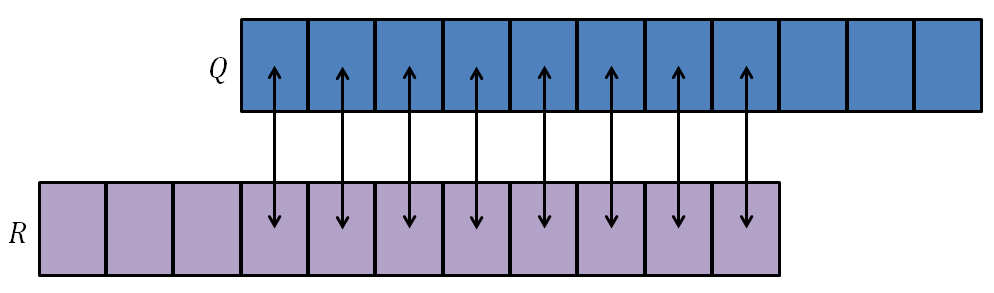}
	\caption{Illustartion of a temporal alignment with $\Delta t' = 3$. In this case we will compute the average error per descriptor with the 8 overlapping frames. The average error is computed by finding NN in both directions, from $V_1$ to $V_2$ and from $V_2$ to $V_1$, and dividing the error by the number of informative descriptors in these frames.}
	\label{fig:integerFrameShift}
\end{figure}

\item \textbf{Computing the sub-frame shift - }since the true $\Delta t$ between the videos is not always an integer value, we search for the sub-frame shift $-1 \leq \alpha \leq 1$ (discretized with gaps $\Delta \alpha = 0.1$). We create a new version of the first video $V'_1$ with the sub-frame shift $\alpha$ by interpolating two consecutive frames as follows: let $p(t)$ be a pixel in frame $t$ then, $p(t+\alpha) = (1-\alpha) p(t) + \alpha p(t+1)$.

We find the informative descriptors $S'_1$ in $V'_1$ and repeat the process of computing the average error per descriptor, but this time with a sub-frame shift of $\Delta t = \Delta t' + \alpha$. Eventually, we choose the sub-frame shift $\Delta t$ with the minimum average error.

\end{enumerate}

\subsection{Spatial Alignment}
Given the estimated temporal alignment, we can proceed to estimate the spatial transformation between the two sequences using these frame correspondences. Let $p_1(t_1) = (x_1,y_1,1)^T$ denote the homogeneous coordinates of only the spatial part of the point $p_1 = (x_1,y_1,t_1)$ in video $V_1$. Similarly, $p_2(t_2) = (x_2,y_2,1)^T$ denote the homogeneous coordinates of the spatial part of its NN $p_2 = (x_2,y_2,t_2)$ in $V_2$ (in frame $t_2 = rt_1 + \delta t$).

We consider two cases: 2D parametric alignment and 3D transformation using epipolar geometry. For each of these cases, the geometric transformation $T_{spatial}$ is a different model and we will describe it in detail in Sec.~\ref{sec:Affine} and \ref{sec:epipolar}.
However, the process of finding the parameters of $T_{spatial}$ is common to both cases. We first find good correspondences (NNs) between corresponding frames across the videos, as described in Sec.~\ref{goodCorrespondences}. We then apply a modified version of RANSAC, using following steps:

\begin{enumerate} \label{spatialTransAlgorithm}
\item Based on the known parameters of the temporal alignment, find good correspondences between corresponding frames in $V_1$ and $V_2$ as described in Sec.~\ref{goodCorrespondences}.

\item Choose at random a subset of pairs of point correspondences.

\item Estimate candidate parameters for $T_{spatial}$ on the selected subset of points.

\item Compute the error score (averaged over all the corresponding descriptors) for the estimated spatial transformation $T_{spatial}$.

\item Repeat steps (2),(3) and (4) N times.

\item Choose the estimated spatial transformation which obtained the lowest error score.

\end{enumerate}

\subsubsection{Affine transformation} \label{sec:Affine}
 When the centers of the cameras are relatively close to each other (compared to their distance to the scene), or when the scene is planner, a 2D parametric transformation suffices to model the spatial transformation between the two video sequences. The most general 2D parametric transformation which models these cases is a 2D projective transformation (a homography). We used a more limited transformation in our algorithm, a 2D affine transformation.
\begin{equation} \label{eq:affineTransform}
\begin{aligned}
& p_2(t_2) = p_2(r t_1 + \Delta t) = A p_1(t_1) 
\\ & \text{ where } A = 
\begin{pmatrix}a_{11} & a_{12} & a_{13}\\ a_{21} & a_{22} & a_{23} \\ 0 & 0& 1 \end{pmatrix}
\end{aligned}
\end{equation}
In this case there are 6 spatial unknown parameters:
\begin{equation*}
T_{spatial} = \begin{bmatrix}a_{11} & a_{12} & a_{13} & a_{21} & a_{22} & a_{23}  \end{bmatrix}
\end{equation*}
For estimating a candidate for the affine transformation $A$ (Step (3)), we need to choose 3 pairs of points at random in Step (2) (since each pair of points contributes 2 equations and there are 6 unknown parameters).

In Step (4) we measure the error of the model in the following way: first we apply the estimated affine transformation on the points in the first video $V_1$ to get their location in the second video $V_2$ as described in Eq.~\ref{eq:affineTransform}. Next, we compute the error between the descriptors in video $V_1$ and the corresponding descriptors in $V_2$. Thus,
\begin{equation}
err(A) = \sum_{p_1(t_1) \in P_1} \Vert d_{V_1}[p_1(t_1)] - d_{V_2}[A p_1(t_1)] \Vert_2
\end{equation}
$P_1$ is the set of the points in $V_1$ whose descriptors are informative, and $A p_1(t_1)$ is their location after applying the affine transformation. $d_{V_1}[\cdot]$ and $d_{V_2}[\cdot]$ denote the descriptors taken from videos $V_1$ and $V_2$, respectively.

We tested our method on several videos (see full videos in \url{http://www.wisdom.weizmann.ac.il/~vision/temporalNeedle/video-alignment.html}). Our temporal needle descriptor consisted of 3x3 patches, a temporal radius of $\Gamma=3$ and 3 temporal scales ($s = 1,\frac{1}{2}, \frac{1}{4}$). We experimented with the following types of videos:
\begin{enumerate}
\item \textbf{Videos with non rigid motion - }Fig.~\ref{fig:flagAlignment} shows an example of temporal and spatial alignment of two videos of flags waving in the wind. 
We first found that the frame shift is -31 frames, which means that the first video $V_1$ is 31 frames behind the second video $V_2$. It can be observed in  (a) that -31 is the temporal shift which obtains the minimum average NN error per descriptor. In this case, the integer frame shift already gave satisfactory temporal alignment. Columns (b) and (c) show frames 207, 211 and 214 in $V_1$ and $V_2$, before temporal and spatial alignments. Column (d) shows the spatial and temporal misalignment in these frames (taking the green band from one video, and the red and blue from the other). Column (e) shows the result after alignment (using the same visualization). The "true color" observed in column (e), indicates that we were able to obtain good alignment between the 2 sequences, both in time and in space.

\begin{figure*}[h!t]
	\centering
	\includegraphics[width=1\textwidth]{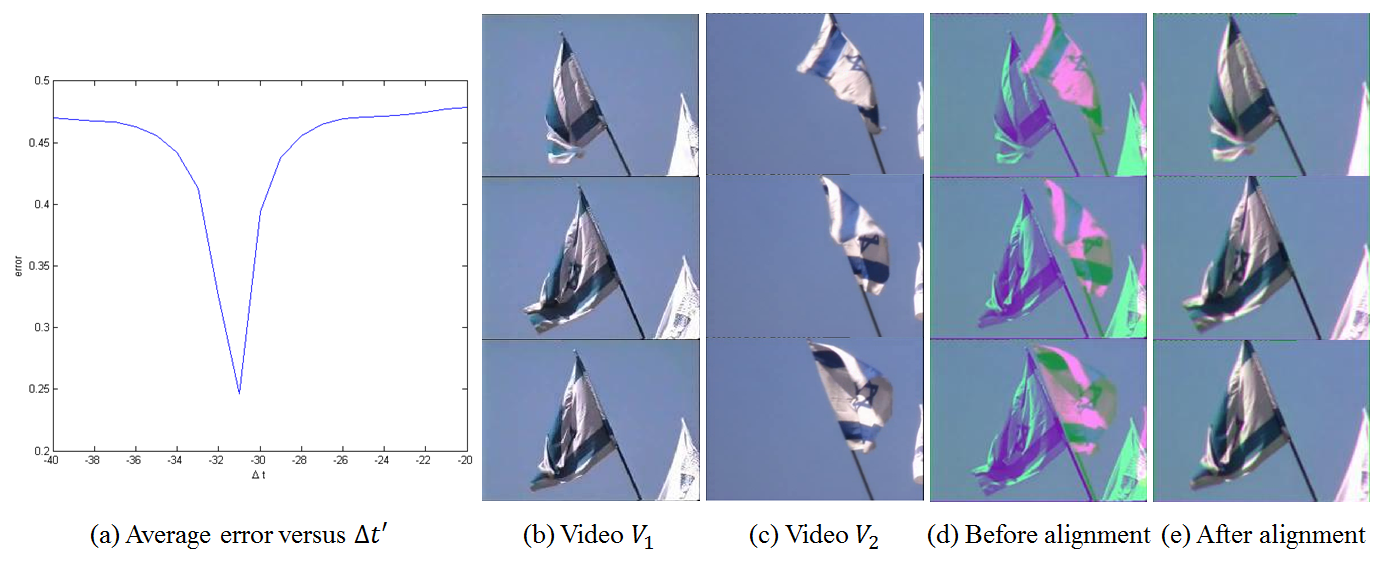}
	\caption{Results of temporal and spatial alignment of non rigid motion. (a) shows the average NN error per descriptor over integer frame shift $\Delta t'$. The minimum average error is obtained for $\Delta t' = -31$, which means that the first video $V_1$ is 31 frames behind the second video $V_2$. Columns (b) and (c) show frames 207, 211 and 214 in $V_1$ and $V_2$, before temporal and spatial alignments. Column (d) shows the spatial and temporal misalignment in these frames (taking the green band from one video, and the red and blue from the other). (e) shows the result after alignment (using the same visualization). The "true color" observed in  indicates that we were able to obtain good alignment between the 2 sequences, both in time and in space}
	\label{fig:flagAlignment}
\end{figure*}

\item \textbf{Videos with significant zoom difference - }Fig.~\ref{fig:zoomAlign} shows an example of aligning two videos with a zoom ratio of 1:3. In this case we recovered only the spatial alignment, since the videos were already synchronized in time. Each row in Fig.~\ref{fig:zoomAlign} shows a different pair of frames. Column (\subref{fig:zoomAll}) shows the results after the spatial alignment (green from one sequence, red and blue from the other). The "true color" obtained in the overlapping regions indicates accurate alignment.

\begin{figure*}[h!t]
	\centering
	\begin{subfigure}[b]{0.3\textwidth}
		\centering
		\includegraphics[width=\textwidth]{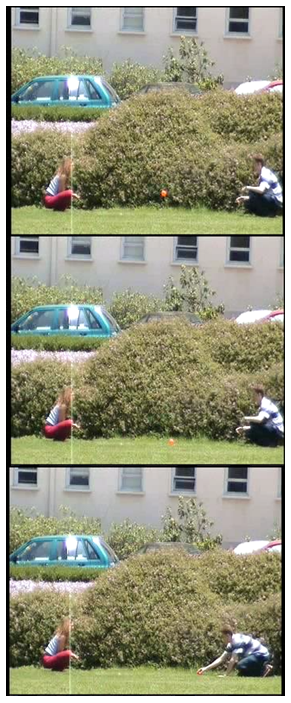}
		\caption{Zoom-out}
		\label{fig:zoomOut}
	\end{subfigure}
	\hfill
	\begin{subfigure}[b]{0.299\textwidth}
		\centering
		\includegraphics[width=\textwidth]{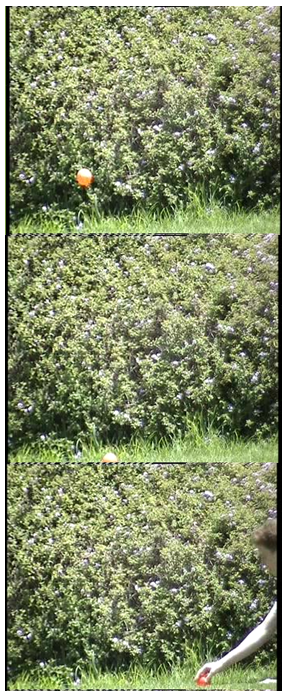}
		\caption{Zoom-in}
		\label{fig:zoomIn}
	\end{subfigure}
	\hfill
	\begin{subfigure}[b]{0.308\textwidth}
		\centering
		\includegraphics[width=\textwidth]{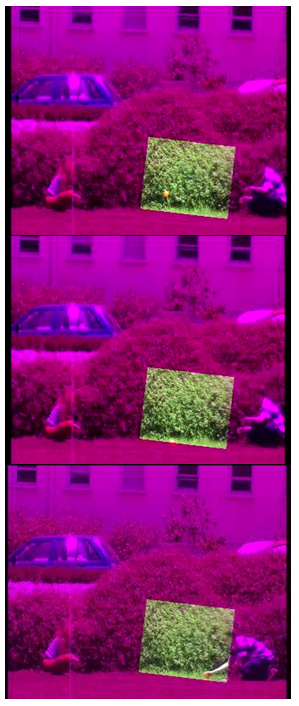}
		\caption{After Alignment}
		\label{fig:zoomAll}
	\end{subfigure}
	\caption{Videos with a large zoom difference of a ball thrown from side to side. We present 3 represantative frames, each row in (\protect\subref{fig:zoomOut}) and (\protect\subref{fig:zoomIn}) present a different pair of frames. (\protect\subref{fig:zoomAll}) shows the results after the alignment (taking the green band from one video, and the red and blue from the other). The "true color" obtained in the overlapping regions indicates accurate alignment.}
	\label{fig:zoomAlign}
\end{figure*}

\item \textbf{Multi Sensor Alignment - } the temporal needle descriptor is appearance invariant. We tested this property by aligning videos that present the same scene with different sensors. We used a short part of a video from Youtube that presents a person walking and the scene is captured with 3 types of sensors (regular daylight camera, camera with night vision device and thermal camera). All 3 multi-sensor videos were successfully aligned by our algorithm. Fig.~\ref{fig:sensors1} shows the alignment results of two of the videos: one is a regular daylight sensor with high gain, and the second is a thermal sensor. Each type of sensor provides different information: in the thermal sensor (Fig.~\ref{fig:sensorsMan1}) we can see more details of the walking person, while in the daylight sensor (Fig.~\ref{fig:sensorsMan2}) the details in the background are clear and the person is visible also through the window, when walking behind it. The result of fusing the two videos is shown in Fig.~\ref{fig:sensorsManFused}, capturing the details from both sequences.

\begin{figure*}[h!t]
	\centering
	\begin{subfigure}[b]{0.3\textwidth} 
		\centering
		\includegraphics[width=\textwidth]{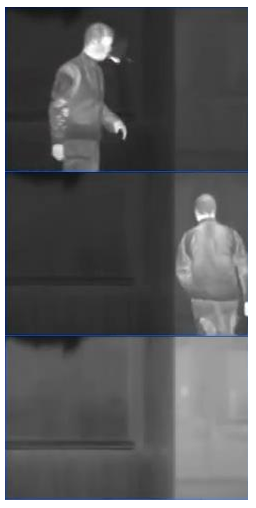}
		\caption{Thermal camera}
		\label{fig:sensorsMan1}
	\end{subfigure}
	\hfill
	\begin{subfigure}[b]{0.302\textwidth} 
		\centering
		\includegraphics[width=\textwidth]{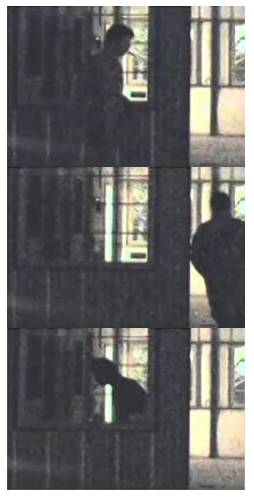}
		\caption{Daylight camera}
		\label{fig:sensorsMan2}
	\end{subfigure}
	\hfill
	\begin{subfigure}[b]{0.32\textwidth} 
		\centering
		\includegraphics[width=\textwidth]{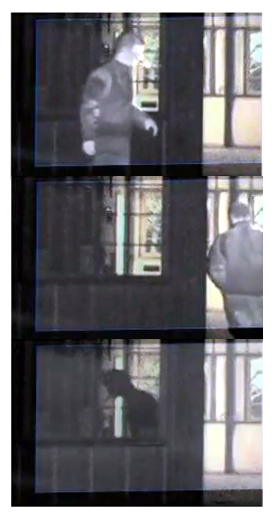}
		\caption{Fused result}
		\label{fig:sensorsManFused}
	\end{subfigure}
	\caption{Multi-sensor alignment. (a) shows 3 frames from the video taken with a thermal video camera. In the last frame we can not see the person since he is behind the window. (b) shows the same 3 frames taken with a regular daylight camera. We can see less details on the person, but the background is clear. (c) shows the fused sequence using the affine transformation that was estimated. It provides the details from both videos.
	}
	\label{fig:sensors1}
\end{figure*}

\item \textbf{Alignment of similar actions (from different scenes) - } Fig.~\ref{fig:manApe} shows an example of a short video taken behind the scenes of "Dawn Of The Planet of The Apes" movie. The first video presents an actor, and the second video is a synthesized video of an ape, that was generated by imitating the movements of the actor. While there is no single global affine transformation between the videos (since the head of the ape was modified differently than its body), we still were able to compute the best affine transformation between the two sequences.

\begin{figure*}[h!t]
	\centering
	\begin{subfigure}[b]{0.235\textwidth}
		\centering
		\includegraphics[width=\textwidth]{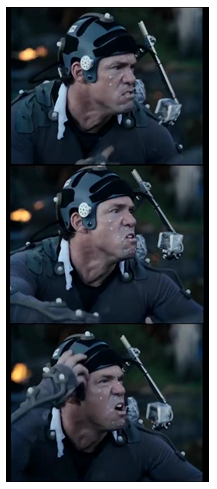}
		\caption{Video $V_1$}
		\label{fig:man}
	\end{subfigure}
	\hfill
	\begin{subfigure}[b]{0.236\textwidth}
		\centering
		\includegraphics[width=\textwidth]{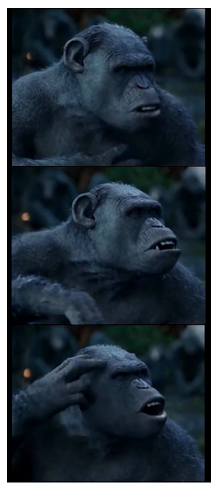}
		\caption{Video $V_2$}
		\label{fig:ape}
	\end{subfigure}
	\hfill
	\begin{subfigure}[b]{0.253\textwidth}
		\centering
		\includegraphics[width=\textwidth]{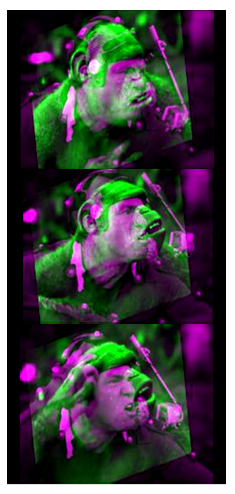}
		\caption{Alignment result}
		\label{fig:manApeAligned}
	\end{subfigure}
	\hfill
	\begin{subfigure}[b]{0.252\textwidth}
		\centering
		\includegraphics[width=\textwidth]{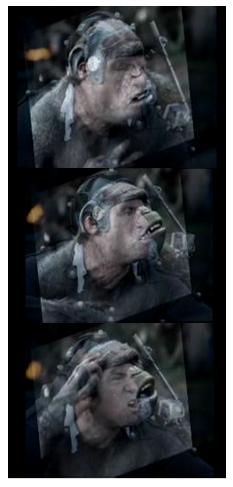}
		\caption{Fused result}
		\label{fig:manApeFused}
	\end{subfigure}
	\caption{Example of spatial alignment between two videos that present the same event. (a) and (b) show 3 different corresponding frames from videos $V_1$ and $V_2$, one of an actor and one of an ape, performing the same action. While there is no single global affine transformation between the videos (since the head of the ape was modified differently than its body), we still were able to compute the best affine transformation between the two sequences. (c) shows the alignment by taking the red and blue bands from the actor's video, and the green band from the ape's video. (d) display the fusion between the videos}
	\label{fig:manApe}
\end{figure*}

\end{enumerate}

\subsubsection{Epipolar Geometry}\label{sec:epipolar}
When the centers of the cameras are located far from each other and the scene is not planar, there is observable parallax between the videos. In this 3D case the spatial relation between the videos is expressed by an unknown 3x3 fundamental matrix $F$:
\begin{equation}
p_2(r t_1 + \Delta t)^T F p_1(t_1) = 0
\end{equation}
In this case there are 9 unknown spatial parameters, the 9 elements of the 3x3 fundamental matrix (although there are fewer degrees of freedom).

We estimate the candidate fundamental matrix $F$ using an implementation of the normalized 8-point algorithm in Hartely and Zisserman \cite{HartleyZisserman:1} (page 281-282).

\begin{figure*}[h!t]
	\centering
	\includegraphics[width=1\textwidth]{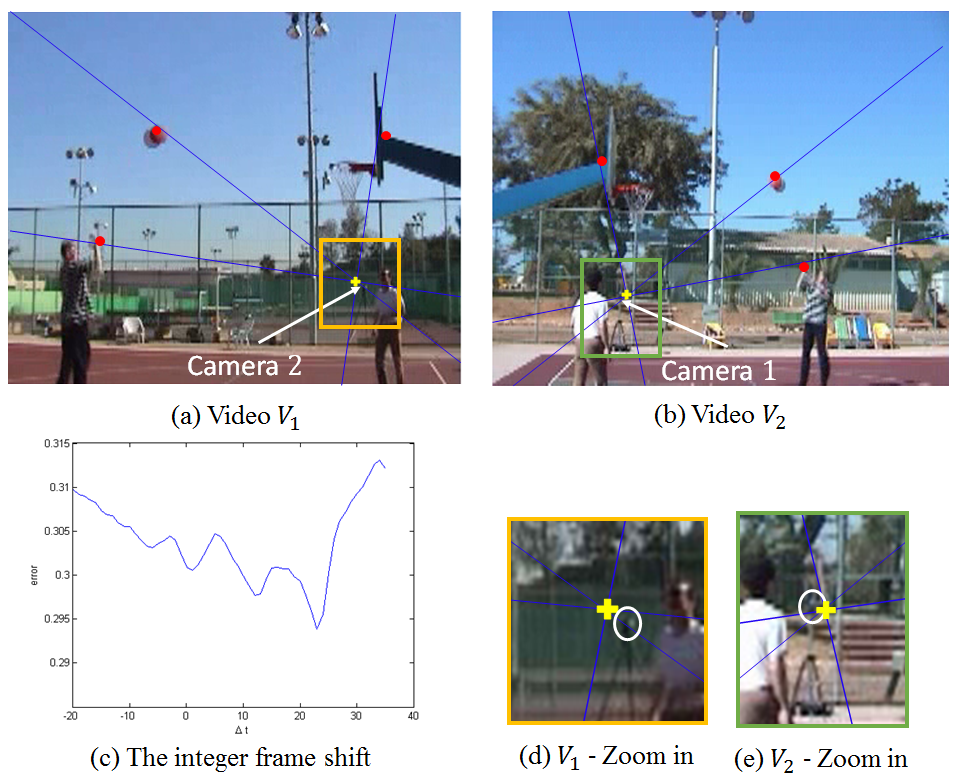}
	\caption{\textbf{An extreme baseline example (cameras facing each other):} temporal alignment and recovering the 3D spatial transformation of videos that capture the same basketball game with a wide baseline. We first aligned the videos temporally, the recovered frame shift was 23.7 frames. (c) display the error over $\Delta t'$ values as a function of integer frame shifts. (a) and (b) display a pair of corresponding frames after the temporal alignment. They show 3 pairs of points and the epipolar lines that corresponds to these points. The epipole that was found is marked by a yellow plus sign, and the true epipole is located at the position of the other camera (which is visible in the frame). (d) and (e) are zoom-in of the regions around the cameras in both videos. As can be seen, the true and the estimated epipoles are very close.}
	\label{fig:basketballEpipolar}
\end{figure*}

To measure the error score for the candidates of the fundamental matrix (step (4) in the Ransac algorithm), we evaluate the first order approximation of the geometric error (Sampson distance) of the fit of a fundamental matrix with respect to a set of matched points as needed by Ransac (Hartely and Zisserman \cite{HartleyZisserman:1} page 287).
\\
\\We tested our method on different examples. Fig.~\ref{fig:basketballEpipolar} shows an example of an extreme wide baseline between the cameras. The videos display a basketball game captured with cameras facing each other. Each camera is visible in the video recorded by the other camera. The recovered temporal shift was $+23.7$ frames. In this extreme case, it is easy to estimate how well  the recovered fundamental matrix is, since the epipole in the first video $V_1$ should fall on the image of the second camera in $V_1$. And vice versa - the epipole of the second video should fall on the image of the first camera in $V_2$. Fig.~\ref{fig:basketballEpipolar} shows that the estimated epipole (marked by a yellow plus sign) is very close to the true epipole (the location of the other camera).

\section{Action Detection} \label{sec:actionDetection}
Let $R(x,y,t)$ be a reference video, and $Q(x,y,t)$ be a template query video  that contains a query dynamic behavior. By action detection we refer to the ability to detect the space-time location of the template $Q$ in video $R$. We would like to find where the action took place in $R$. The query and the reference videos do not have to be of the same spatial size or temporal length.

Detecting an action is conceptually similar to aligning videos both in space and time. We can detect an action by finding good correspondences between statistically significant descriptors in the query video, to the descriptors in the reference video.
Unlike the alignment case, in this case we are only interested in finding good correspondences in a smaller space-time region, for the desired action, and not in the entire video. Moreover, the match may occur at multiple positions in the reference video (e.g., if repeated several times).

\begin{figure*}[h!t]
	\centering
	\includegraphics[width=1\textwidth]{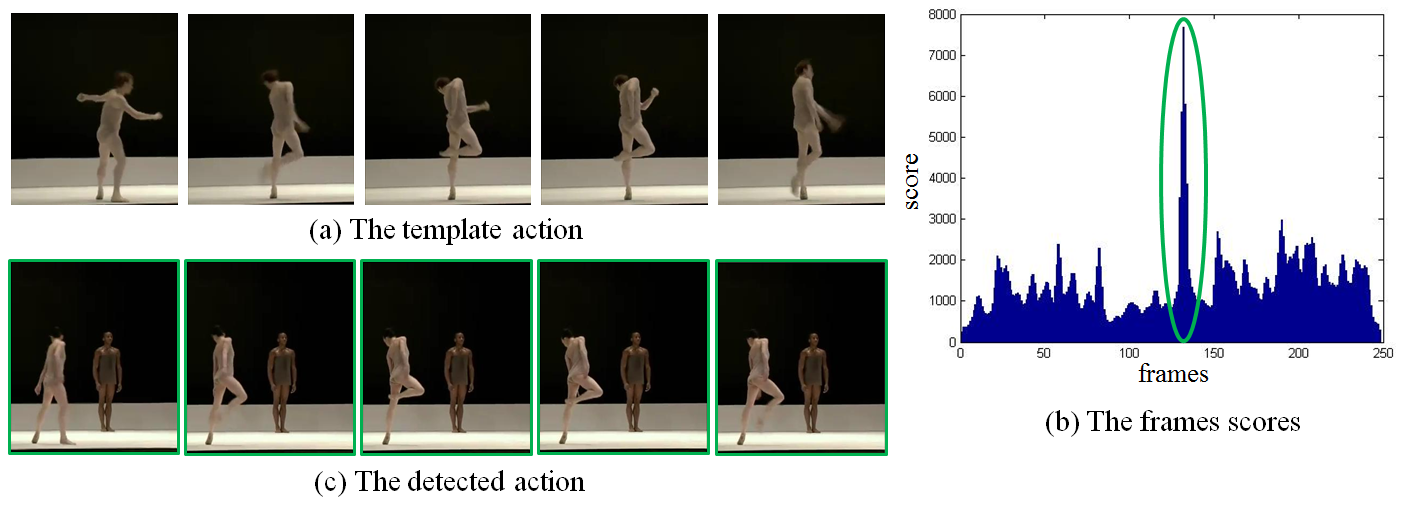}
	\caption{\textbf{Action detection in dance video.} We selected a single dance move performed by a male dancer as a template. (a) shows a few frames from the action. (b) shows the score assigned to every frame in the reference video, signifying how likely it is to be the center-frame of the detected action. Our algorithm correctly detected the action, with no false alarms and no mis-detections. (c) shows the corresponding frames of the action that was detected}
	\label{fig:balletActionDetection}
\end{figure*}

\begin{figure*}[h!t]
	\centering
	\includegraphics[width=1\textwidth]{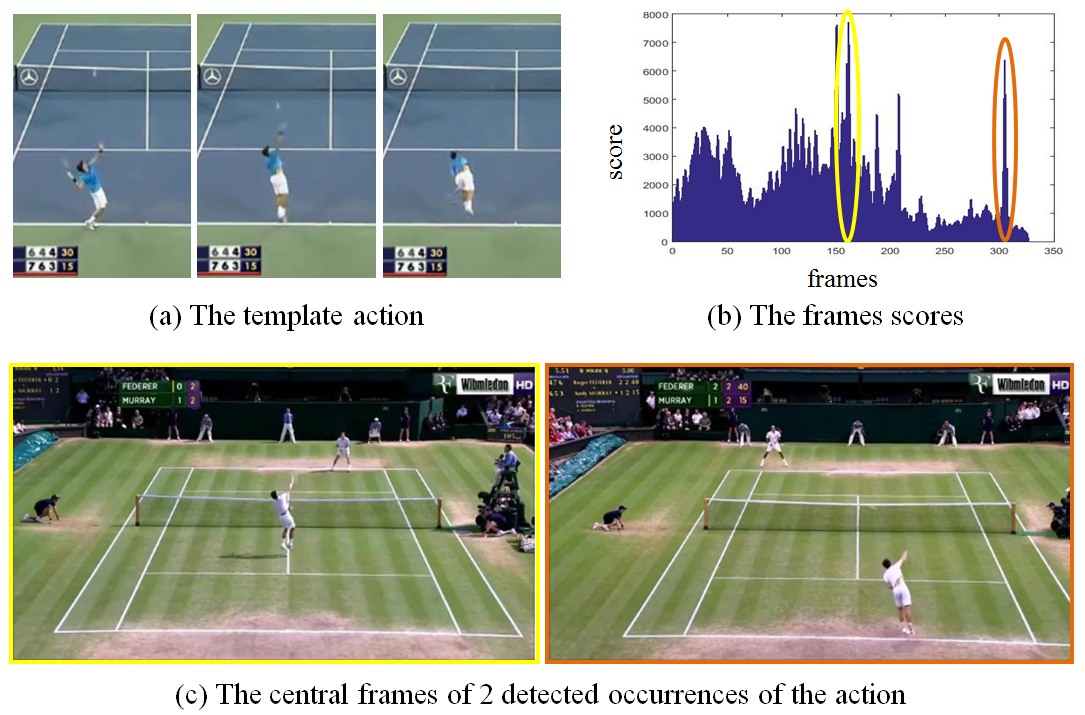}
	\caption{\textbf{Action detection in sports video.} We selected a tennis serve as an action template. (a)~shows three frames from the action. (b)~shows the score assigned to every frame in the reference video, signifying how likely it is to be the center-frame of the detected action. (c)~shows the center frame of two detected occurrences of the action. Our algorithm correctly detected the actions, with no false alarms and no mis-detections.}
	\label{fig:tennisActionDetection}
\end{figure*}

The process of action detection is done as follows:
\begin{enumerate}
\item We find the informative descriptors in the template video $Q$ as described in Sec.~\ref{goodCorrespondences}.

\item For each informative descriptor $d$ in the action template, we search for 15 NNs (nearest neighbors) in video $R$. Each NN of $d$ votes to a frame in $R$ as a candidate center frame of the action, the same way as $d$ relates to the center frame of the query $Q$. For example, suppose an informative descriptor $d\in Q$ is $\alpha$ frames after the center frame of $Q$, then every one of its NNs will vote to $\alpha$ frames before their frame in $R$ as a candidate central frame of the detected action.

\item We define a frame as a detected action center, if its score provides a local maximum with value larger than twice the average frame-score. If an action occurs more than once, we are able to detect all of its occurrences. In all cases we experimented with, when the action was not present at all, the values of all frames were almost the same (none exceeded twice the average score) and no frame was be detected. However, we can envision cases where there will be false alarms, in these cases our method will detect the most similar action.

\item Once we detected the center frame, we can display the correspondences that were found. We can further run spatial alignment as described in Sec.~\ref{sec:Affine}, to find the affine transformation which maximizes the similarity between the template and the reference video (it only applies when the actions in $Q$ and $R$ were taken from similar 3D viewpoints).

\end{enumerate}
We tested our method on several videos, in some cases we chose the query to be one instance of an action that occurs multiple times in the reference video. Fig.~\ref{fig:balletActionDetection} and \ref{fig:tennisActionDetection} show two examples display the results of two experiments. Their videos as well as additional examples can be found in \url{http://www.wisdom.weizmann.ac.il/~vision/temporalNeedle/action-detection.html}.
\\
\textbf{Experiment with videos of a modern ballet - } we chose as a query $Q$ a short video segment of a male dancer performing a specific move. Fig.~\ref{fig:balletActionDetection} (a) shows few frames from the move. The reference video $R$ is a video of a female dancer that contains many moves. In our experiment we detected the action correctly with no false alarms, as can be seen in Fig.~\ref{fig:balletActionDetection}(b). Fig.~\ref{fig:balletActionDetection}(c) shows the corresponding frames of the action that was detected. The two dancers are supposed to perform the same dance, but in fact there are some small variations in their performance. For example, the position and height of the arms are different between the dancers. Also, the view point of the camera relative to the dancer is not identical for the two dancers. The algorithm overcame these small variations and is able to detect that the dancers perform similar moves.
\\\textbf{Experiment with videos of tennis games - } Fig.~\ref{fig:tennisActionDetection} display the results of detecting a tennis serve in two different games. We chose as a query a serve from one tennis game, (a) shows three frames from the query action. The reference video $R$ is a longer segment from a different tennis game with different players. (b) shows the score of every frame in the reference video to be an action center. Our algorithm detected the action twice (marked by yellow and orange). (c) shows the center frame of the actions that were detected. In tennis, serve and smash hits are very similar, therefore we refer to them as the same action. In this example the first detection (with yellow frame) detected a smash and the second detection detected a serve. The algorithm detected the actions correctly and with no false alarm and mis-detection.

\section{Video Clustering} \label{sec:videoClustering}
"Clustering by Composition" \cite{Michal:2} partitions a collection of images into clusters of similar image categories by the affinities between the images in the collection. The affinity between two images is computed based on finding large non-trivial shared regions between the images.

In our work we extend the algorithm to videos. Combining it with the temporal needle descriptor, we are able to discover automatically categories from a collection of unlabeled videos. Our method contains two steps: (1) Building an affinity matrix to reflect the videos relations based on shared space-time volumes between the videos. The affinity between videos build on top of our temporal needle. (2) Partition the videos into clusters based on the affinity matrix using N-cuts \cite{Ncut:1}.

The main key to building the affinity matrix is the \textbf{region growing algorithm}. We slightly modified the region growing of \cite{Michal:2} to space-time, to handle videos instead of images. For full details and proofs we refer the reader to "Clustering by Composition" \cite{Michal:2} Sec.4. 

Let $R$ be a 3D (space-time) shared region (with unknown size and shape) between videos $V_1$ and $V_2$. Let $F$ denote the number of frames in $V_1$ and $V_2$, and $N$ the number of pixels per frame. Denote by $R_1$ and $R_2$ its instances in $V_1$ and $V_2$, respectively. The goal of the algorithm is to find for each descriptor $d_1 \in R_1$ its matching descriptor $d_2 \in R_2$. 

The algorithm is composed by two steps: 
\\\textbf{The sampling step - } in this step, every descriptor $d_1 \in V_1$ randomly samples $S$ positions $(x,y,t)$ in $V_2$ and chooses its best matching descriptor among them. The run time of this step is $O(SNF)$.
\\\textbf{The propagation step - } each descriptor chooses between its best match in the sampling step and matches proposed to it by its spatio-temporal neighbors. This is achieved by sweeping four times the video: two spatial sweeps (for each frame once from top down and the second from bottom up) and two temporal sweeps (once from the beginning to end and once from the end to the beginning). The run time of this step is $O(NF)$.
\\\textbf{Time complexity - } the overall running time of the algorithm: $O(SNF)$, namely, linear in the size of the video. 
According to the theory in \cite{Michal:2}, in order to detect a region of size $|R|$ with probability $p \geq (1-\delta)$, the required number of samples is $S = \frac{NF}{|R|} log (\frac{1}{\delta})$. 
\\For example, if we assume that the shared space-time region of the action is 10\% of the spatial size and 10\% of the temporal size, it results with a shared region of size 1\% of the entire video. Hence, for $\delta = 2\%$ (probability of detection $p\geq98\%$) the number of samples required to detect this region is constant, $S = 392$.
\\
\\After we modified the region growing algorithm, we used the same algorithm for finding the \textbf{collaborative video clustering}. For full details we refer to "Clustering by Composition" \cite{Michal:2} Sec.5. We briefly describe the algorithm bellow: 
\\\underline{Clustering Algorithm:} We start with a uniform random sampling.
Each descriptor randomly samples $S=392C$ samples across the videos in the collection. Where $C$ is the number of clusters. At each iteration, using the region growing algorithm, shared space-time regions are found, inducing connections between videos in the collection. These connections induce a sparse set of affinities between the videos. The sampling density distribution of each video is updated according to the affinities. For example, if video $V_1$ has high affinity with videos $V_2$ and $V_3$, in the next iteration videos $V_2$ and $V_3$ will be encouraged to sample more in each other. This results in a "guided" sampling, exploiting the "wisdom of crowds of videos". Finally, after several such iterations, the resulting affinities are fed to N-cut algorithm \cite{Ncut:1}, to partition the videos into the desired $C$ clusters.
\\\textbf{Time and memory complexity - } Let $M$ be the number of videos in the collection and $T$ be the number of iteration. Each iteration runs the region growing algorithm for every video and re-estimate the sampling density distribution for the next iteration. Therefore, each iteration takes $O(MNF)$ and overall the algorithm takes $O(TMNF)$. However, since the affinity matrix should be sparse, the number of iterations is typically small, $T = 10 log_{10} M$, thus the runtime is $O(MNFlogM)$.
\\During the algorithm computation, we hold all the descriptors of the videos in the collection in the memory. Therefore, the memory is $O(MNF)$.
\\
\\The time and memory complexity leads to the main limitation of our algorithm for video clustering. The memory and run time are both proportional to the number of videos in the collection and their size. This restricted us to using small collections of short videos in our experiments.

We tested our algorithm on two collections of videos that we created. Most of the videos were taken from Youtube, and some were taken from UCF-Sports dataset \cite{UCF}. We downloaded videos from Youtube because most of the action recognition datasets contain very short videos that focus on distinguishing  single and simple actions. We were interested in using longer and more complex videos, so we can find interesting common space-time regions across videos in the same category.

\begin{enumerate} [(a)]
\item Judo and Karate collection - we collected 14 Karate and 14 Judo videos from Youtube. These two martial arts have similar spatial appearances, and vary in the specific unique movements. The unique movements are usually detected as the shared space-time regions between videos in the same cluster. The videos spatial frame size is 360x480 and their length varies between 2 to 5 seconds. We ran our algorithm with 5 iterations only and it was able to assign correctly 25/28 videos, results with $89.3\%$ mean purity. The results are presented in Fig \ref{fig:karateJudo}.

\begin{figure*}[h!t]
	\centering
	\includegraphics[width=1\textwidth]{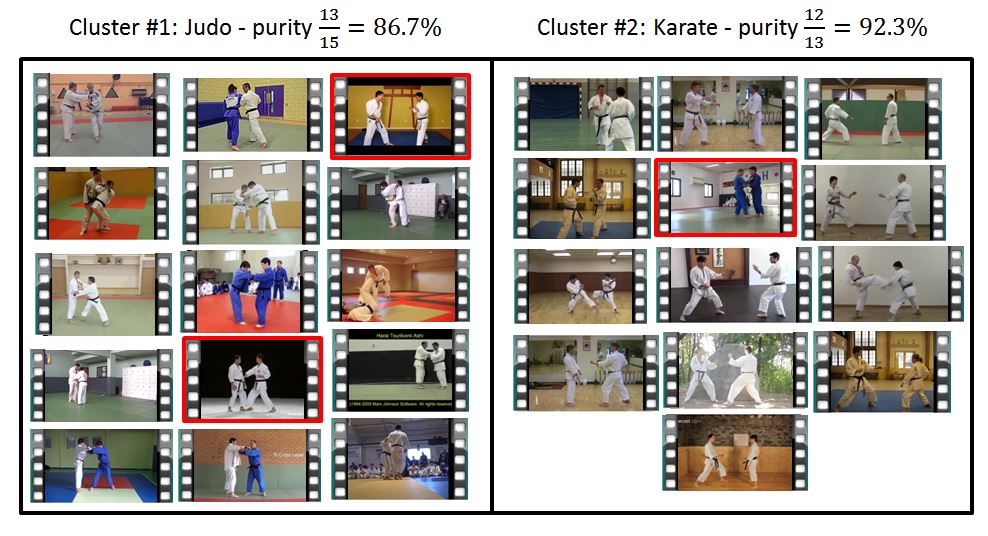}
	\caption{Clustering results on a small collection of Karate and Judo videos we collected from Youtube. The collection contains 14 Katrate and 14 Judo videos, they are all presented in the figure. Videos that were assigned to the wrong cluster are marked by a red rectangle. We assigned correctly 25 out of 28 videos, which results with mean purity of 89.3\%}
	\label{fig:karateJudo}
\end{figure*}

\item Skateboarding and Walking collection - the collection includes 15 skateboarding and 15 walking videos. Some of the videos were taken from UCF-sports dataset \cite{UCF}, but most of them were downloaded from Youtube. The spatial frame size varies between the videos, and their length is between 2 to 5 seconds. Fig.~\ref{fig:SkateboardWalk} display the results, the algorithm correctly clustered 27 out of 30 videos, which results with mean purity of $90\%$.

\begin{figure*}[h!t]
	\centering
	\includegraphics[width=1\textwidth]{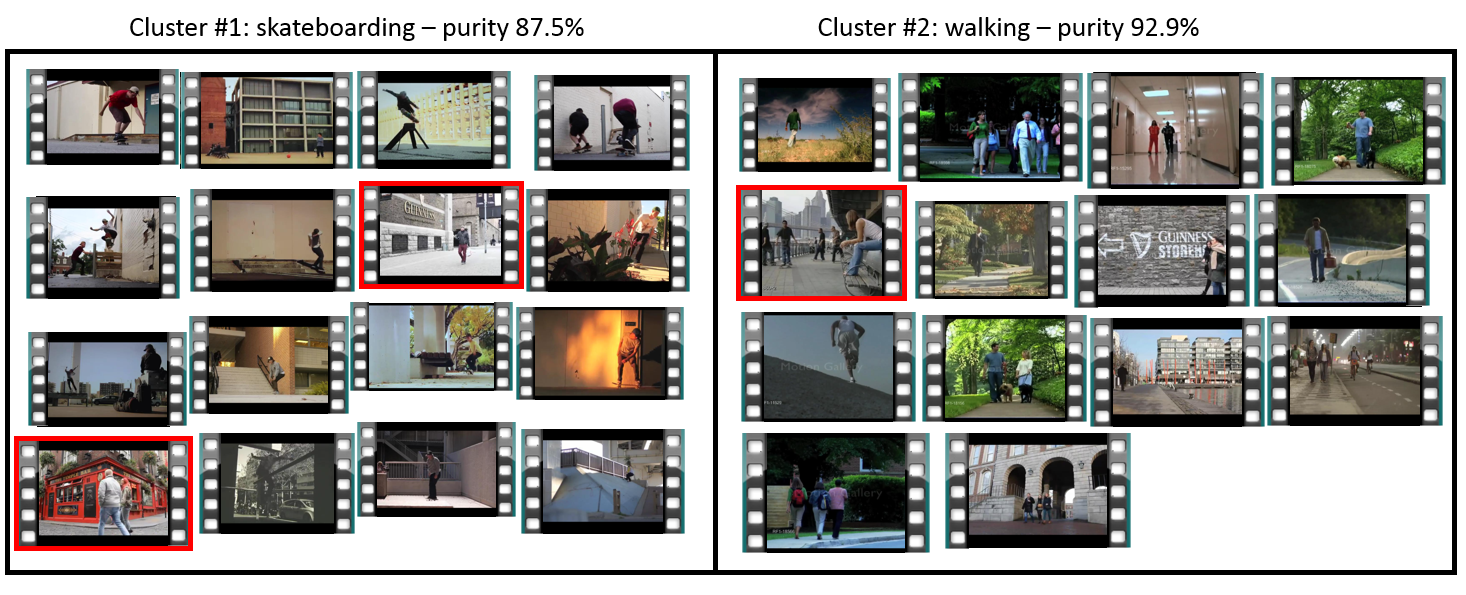}
	\caption{Clustering results on a small collection of skateboarding and walking videos we collected from Youtube and UCF-Sports dataset. The collection contains 30 videos, 15 skateboarding and 15 walking videos. One frame of each video is presented in the figure. The videos assigned to the wrong cluster are marked by a red rectangel. The algorithm assigned 27 out of 30 videos correctly which results in $90\%$ mean purity. }
	\label{fig:SkateboardWalk}
\end{figure*}

\end{enumerate}

\section{Summary} \label{sec:summary}
In this paper we presented the "Temporal-Needle" - a video descriptor which captures dynamic behavior, while being invariant both to appearance and to viewpoint. We showed how using this descriptor gives rise to detection of the same dynamic behavior across videos in a variety of scenarios. In particular, we demonstrated the use of the descriptor in tasks such as sequence-to-sequence alignment under complex conditions, action detection, as well as video clustering for unsupervised discovery of video categories.

\bibliography{temporal_needle_arxiv}
\bibliographystyle{abbrv}

\end{document}